\pdfoutput=1

\documentclass[11pt]{article}

\usepackage{EMNLP2023}

\usepackage{times}
\usepackage{latexsym}

\usepackage[T1]{fontenc}

\usepackage[utf8]{inputenc}

\usepackage{microtype}

\usepackage{inconsolata}

\usepackage{enumitem}
\usepackage{amsfonts}
\usepackage{amsmath}  
\usepackage{amsthm}
\usepackage{multirow}
\usepackage{algorithm}
\usepackage[many]{tcolorbox}

\usepackage{subcaption}
\usepackage{booktabs}
\usepackage[textsize=scriptsize]{todonotes}

\usepackage{array}
\usepackage{placeins}

\usepackage{xr}

\newcommand{\nam}{\textsc{mh-block}}
\newcommand{\mnm}{\textsc{m\&m}}

\title{A Block Metropolis-Hastings Sampler \\for Controllable Energy-based Text Generation}

\author{Jarad Forristal${}^1$ \\ \texttt{jforristal@ucsd.edu} \\  \And 
        Niloofar Mireshghallah${}^1$ \\ \texttt{fmireshg@ucsd.edu} \\
        \AND
        Greg Durrett${}^2$  \\ \texttt{gdurrett@cs.utexas.edu} \\  \And
        Taylor Berg-Kirkpatrick${}^1$ \\ \texttt{tberg@ucsd.edu} \\ \AND  
        \\ ${}^1$Department of Computer Science and Engineering, The University of California San Diego \\ ${}^2$Department of Computer Science, The University of Texas at Austin}

\begin{document}
\maketitle

\begin{abstract}
Recent work has shown that energy-based language modeling is an effective framework for controllable text generation because it enables flexible integration of arbitrary discriminators. However, because energy-based LMs are globally normalized, approximate techniques like Metropolis-Hastings (MH) are required for inference. Past work has largely explored simple proposal distributions that modify a single token at a time, like in Gibbs sampling. In this paper, we develop a novel MH sampler that, in contrast, \textit{proposes re-writes of the entire sequence in each step} via iterative prompting of a large language model. Our new sampler (a) allows for more efficient and accurate sampling from a target distribution and (b) allows generation length to be determined through the sampling procedure rather than fixed in advance, as past work has required.  We perform experiments on two controlled generation tasks, showing both downstream performance gains and more accurate target distribution sampling in comparison with single-token proposal techniques. 

\end{abstract}

\section{Introduction}

Controllable text generation has many important downstream applications, ranging from reducing bias in generated text to increasing factuality~\citep{xudetoxifying,gehman2020realtoxicityprompts,sap2021annotators,baheti2021just,mireshghallah-berg-kirkpatrick-2021-style}. While traditional autoregressive language models (LMs) can produce highly fluent text, controlling their output and generating text which satisfies specific desired attributes remains a hard problem for all but the largest industrial LMs. One line of past work has made progress on controllable text generation by integrating discriminators---e.g.~pretrained text classifiers that directly measure control attributes---into the scoring function for text generation \cite{mnm, fudge, pplm, gedi}. These techniques provide a flexible interface for exerting control: a user can combine discriminators and heuristic scoring functions together with likelihoods from traditional LMs to form a product of experts, guiding outputs to satisfy target criteria. 


\begin{figure*}[!htpb]
\centering 
\includegraphics[width=1\textwidth]{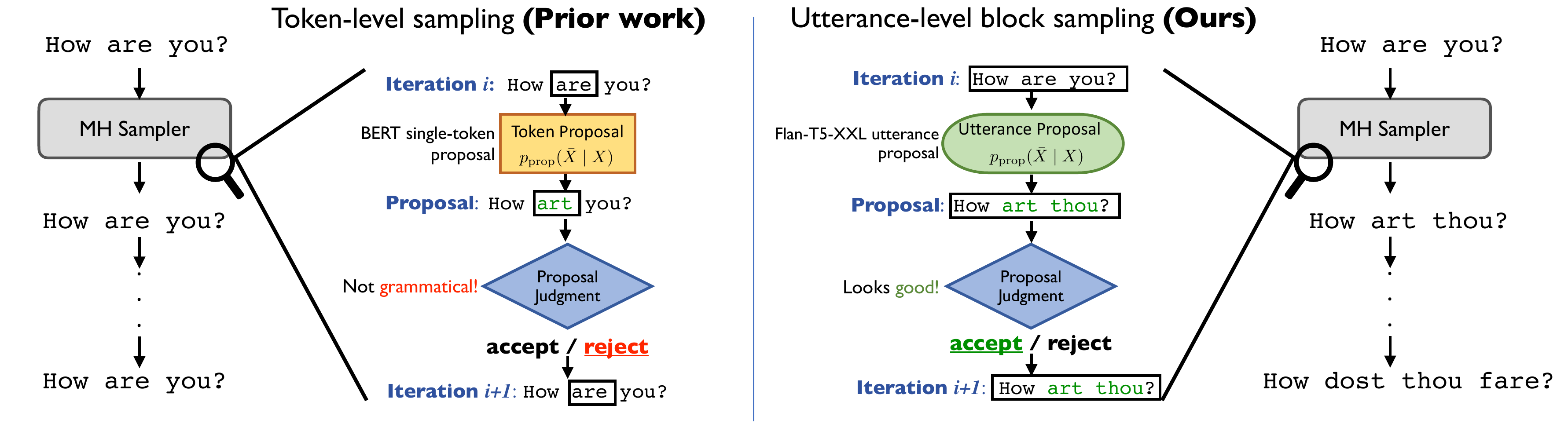}
\vspace{2mm}
\caption{An overview of our novel Metropolis-Hasting (MH) sampler for energy LMs, detailing the iterative editing procedure. For our method, we prompt Flan-T5-XXL to edit the sentence in the desired way, and use this conditional distribution as the proposal for the MH chain. The MH accept/reject step corrects the bias of the proposal by considering the unnormalized energy under the target distribution. If we accept the edit, it becomes the input to Flan-T5-XXL in the next step of the Markov chain. The baseline, in contrast, only propose a change to a single token at a time.}\label{fig:contrast}
\end{figure*}

While these techniques enable effective control, they present a new challenge for decoding. The scoring functions introduced by discriminators are not autoregressive: they are global potential functions that take the entire utterance as input. This means that the overall model is not autoregressive and exact sampling is intractable. Past work has developed various heuristic or approximate decoding strategies \cite{pplm,gedi,fudge,goyal2022exposing,mnm,qin2022cold,kumar-etal-2022-gradient,kumar2021controlled}.
One of the more principled inference techniques treats the product of experts as an energy-based LM---that is, a globally normalized language model~\cite{goyal2022exposing,mnm,qin2022cold,pmlr-v48-belanger16}---and introduces a Metropolis-Hastings (MH) sampler for decoding. More specifically, \citet{mnm} use BERT \cite{bert} to propose a change to a single token of the current sequence at each step of the MH chain (like a traditional Gibbs sampler) and the energy LM exerts its influence through MH's accept/reject step, correcting the bias of the proposal distribution. While principled, this approach has serious limitations. First, since only a single token can be changed at each step, inference is extremely slow. Second, since the proposal distribution does not alter the length of the current sequence, the length of the desired output must be specified in advance. 

In this work, we present a novel MH sampler for energy LMs that, in contrast with past work, introduces a proposal distribution that allows for arbitrary re-writes of the entire sequence at each step of the MH chain. As a result, our \textit{block MH sampler} (a) has improved efficiency in sampling and (b) allows output length to be determined by the sampling process itself. 
Our key insight is to use a prompted large language model (LLM) as the proposal distribution inside of our sampler. Specifically, we prompt the LLM to paraphrase the text sequence at the current step of the MH chain, and use its output distribution as the proposal for the next step.
Whether or not the proposal is \emph{accepted} is still governed by the energy function of the target energy LM; we only change the proposal, while leaving the mathematical framework intact. 

We conduct experiments on two downstream text style transfer tasks that have been used in past work as benchmarks for controllable generation \cite{mnm, strap}. Specifically, we study style transfer performance on two challenging datasets: the Shakespeare author imitation dataset~\cite{shakespeare} and the GYAFC formality corpus~\cite{gyafc}. Across experiments, we find that our novel sampler is able to make substantially faster progress towards high-scoring samples per forward-pass of the target energy LM in comparison with the single-token re-sampling MH procedure from past work. Further, for most downstream tasks, our novel sampler also leads to improvements in the output text in terms of fluency, style transfer accuracy, and semantic similarity to the desired ground truth generations.

\paragraph{Our contributions:} (1) We propose novel block MH sampler for globally normalized energy LMs that is capable of rapid substantive edits; (2) We validate our approach on two downstream controllable generation tasks, formality transfer and author imitation, demonstrating gains in sampling efficiency as well is in output text quality; (3) We conduct an intrinsic evaluation of our sampling procedure in a synthetic setting, comparing outputs from our sampler with outputs from exact ancestral sampling.

\section{Background}


The \mnm{} approach~\cite{mnm} defines an MCMC sampling procedure for language models that are globally normalized, which are often called energy-based LMs. Explicitly, an energy-based sequence model defines a globally normalized probability distribution over the space of possible finite-length sequences $\mathcal{X}$ as: $p(X;\theta) = \frac{e^{-E(X;\theta)}}{\sum_{X' \in \mathcal{X}} e^{-E(X'; \theta)}}$, where $E(X;\theta)$ corresponds to the scalar energy of a sequence $X$ that is judged by some model parameterized by $\theta$. Lower energy corresponds to higher likelihood of $X$. Unlike popular autoregressive techniques, there is no general tractable method of sampling from energy models formulated in this way -- even the likelihood function is intractable to compute due to the global normalization constant. However, their high flexibility and compatibility with black-box experts make energy models highly attractive, warranting research into this problem.

\paragraph{Product of Experts}\label{sec:experts}

The constraints associated with controlled generation can be thought of as distributing probability mass over a small subspace of $\mathcal{X}$ associated with samples that satisfy the required constraints. For example, if we want to generate Shakespearean sentences, we likely want both fluent and early-modern English outputs (modeled by $p_{\text{shakespeare}}(X)$ and $p_{\text{fluent}}(X)$ respectively) -- i.e., $p_{\text{desire}}(X) \propto p_{\text{shakespeare}}(X) \cdot p_{\text{fluent}}(X)$. Because it is intractable to form these probability distributions explicitly, we instead model them implicitly using unnormalized potential functions, combining them to form a scalar energy:

\begin{equation}\label{eq:product}
E(X) = \sum_{i=1}^k \alpha_i E_{i}(X),
\end{equation}
where $a_i$ are scalar weights and $E_i(X)$ are arbitrary black-box potential functions. More information regarding our use of energy models is available in Section~\ref{sec:method} and Section~\ref{sec:fac}.

\paragraph{Sampling from $\mathbf{E(X;\boldsymbol\theta)}$} \mnm{} uses a Metropolis-Hastings (MH) chain with a Gibbs-inspired proposal distribution to sample from the target energy model $E(X;\theta)$. Starting with some text, $X$, for each iteration \mnm{} randomly samples the position of a single token to mask out. BERT is used to propose a new token for the masked position, editing the sentence into $\bar{X}$. This proposed edit, $\bar{X}$ is then accepted or rejected based on the conditional probability of the proposed token, likelihood of the replaced token, and the ratio of energies between $\bar{X}$ and $X$; the exact calculation can be seen in Equation~\ref{eq:acc}. Critically, the energy model's likelihood only appears in the ratio in Equation~\ref{eq:acc} and the \textit{intractable normalization constant cancels out}; this is one of the primary motivations for using MH in this context. The model used to estimate $p(X \vert \bar{X})$ and $p(\bar{X} \vert X)$ is called the \emph{proposal distribution}. The stationary distribution of this Markov chain converges to $p(X;\theta)$. 




\section{Methodology}\label{sec:method}


In this section, we will describe and motivate our approach. Similar to \mnm{}, we frame controlled generation as a sampling problem where our goal is to get samples from a specific energy-based sequence model. However, \mnm{} has important limitations in the sampling procedure that should be noted:

\paragraph{Limitations of Token-level Sampling} The \mnm{} masking process destroys important information that is often relevant to the task at hand: for example, if a name is masked out, it is unlikely to be predicted again; this means \mnm{} can largely not restructure sentences and instead prefers minimal edits which achieve the end goal. Importantly, editing a single token at a time also significantly slows mixing. For example, if we want to make the sentence ``How are you?'' to be more Shakespearean, the single-token edit ``How art you?'' is not fluent or grammatical and is likely to be rejected, but is a necessary step to achieve the end goal of ``How art thou?''; this important issue is illustrated in Figure~\ref{fig:contrast}. Using a block MH sampler sidesteps this issue by allowing the proposal distribution to select which parts of the sentence to edit and to propose changes to multiple tokens simultaneously. 

Furthermore, \mnm{} uses BERT to calculate $p(X \vert \bar{X})$ and $p(\bar{X} \vert X)$. Importantly, since BERT was trained on a dataset of modern English, samples from this distribution will also be. In Figure~\ref{fig:contrast} BERT is unlikely to propose the token ``art'' in the first place, this is not a modern English token and BERT has no information about the task. Prompting an LLM with information about the task guides the model towards making more impactful changes. Finally, \mnm{} is a fixed-length sampling method: the output is always the same length as the input. The freedom to add or delete tokens is very valuable for many downstream tasks. Our sampling procedure, detailed below, targets these weak-points and improves upon past work.

\subsection{Sampling Scheme}
Similar to \citet{mnm}, we devise a Metropolis-Hastings (MH) chain that iteratively edits text in order to produce samples from the target energy model. 
We begin with a set seed text and progressively edit this sentence, forming a long Markov chain in the process. The acceptance or rejection of these edits is a function of both the expert blackbox models and sample probability as judged by the proposal model. Unlike previous work where the proposal function was replacement of a single token, we instead choose to prompt Flan-T5-XXL~\citep{flanxxl} to edit the sentence; this allows for arbitrary-length generation and makes our approach a block-level MH sampler (similar to blocked Gibbs sampling) as multiple variables (tokens) are updated every proposal step.

More specifically, at each step of the chain, given the current sentence $X$, an edited version, $\bar{X}$, is sampled from the proposal distribution, $p_{\textrm{prop}}(\bar X \mid X)$, which is defined by an instance of Flan-T5-XXL that has been prompted to generate paraphrases as depicted in Figure~\ref{prompt}. MH then defines the probability of transitioning from $X$ to $\bar{X}$ as:
\begin{equation}
p(\bar{X};X) = \min\left( 1,\frac{e^{-E(\bar{X})}~p_{\textrm{prop}}(X \mid \bar{X})}{e^{-E(X)}~p_{\textrm{prop}}(\bar{X} \mid X)} \right) \label{eq:acc}
\end{equation} 
$E(X)$ refers to the product of experts energy defined in Equation~\ref{eq:product} and $p_{\textrm{prop}}(\bar{X} \mid X)$ refers to the probability that the proposal model generates $\bar{X}$ given its prompted input is $X$. 

Strictly speaking, to inherit the asymptotic guarantees of MH, one would need to prove, for example, detailed balance conditions for the proposal distribution. However, in practice, we found Flan-T5-XXL to have a strong propensity to generate the identity edit which causes slow mixing. To mitigate this issue in our experiments, in the numerator of Equation~\ref{eq:acc} we instead use $p_{\textrm{prop}}(X \mid X)$. This change makes non-identity edits more likely to be accepted if the probability of the identity is high. In practice, we found this approximate accept/reject strategy to perform well in experiments.   

Thus, our block-level MH sampler implements a more freeform style of editing compared to token-level replacement used in previous work, as illustrated in Figure~\ref{fig:contrast}. Specifically, the block-level sampler: (1) allows the chain to preserve the content of the previous sentence more easily, as we do not mask out or destroy any information, (2) allows for coordinated edits to multiple tokens simultaneously, and (3) allows for the length of the sentence to change over the course of the sampling process. 

In our implementation, we progressively edit a sentence by iteratively reprompting an LLM and accepting or rejecting these edits based on the `quality' of the edit as judged by both the LLM itself and expert black-box models. Rather than running a single Markov chain at a time, we instead opt to run a batch of independent Markov chains with the same initial seed text, selecting a single final generation by selecting the one with minimum energy. We refer to this as ``batch-size'' when describing our experiments; we use batch-size 10 for all experiments unless noted otherwise. Using the methodology now defined, we can leverage the power of LLMs to sample from any arbitrary distribution that can be formulated as an unnormalized energy. 


\section{Intrinsic Evaluation of Sampler}

In this section, we aim to conduct an intrinsic evaluation of the proposed sampler, which we refer to as \nam{}, separate from the downstream controllable generation tasks we consider in Section~\ref{sec:extern_results}. Specifically, we would like to evaluate how well \nam{} approximates exact sampling from a complex target distribution relative to the baseline token-level sampling procedure, which we refer to as \mnm{}. To accomplish this, we need to define a target energy model for which exact sampling is actually tractable so that we can draw exact samples and compare. For this purpose, we treat a prompted conditional distribution of LLaMA-7B \cite{llama_paper} as our target `energy' model by setting $E(X)$ in Equation~\ref{eq:product} to LLaMA-7B's negative log-likelihood. Specifically, we prompt LLaMA-7B to paraphrase a fixed input sentence (randomly sampled from the Shakespeare dataset mentioned in Section~\ref{sec:extern_results}, consisting of 13 tokens) and treat the resulting conditional over text sequences as our target. 

\begin{figure}[t!]
\includegraphics[width=.485\textwidth]{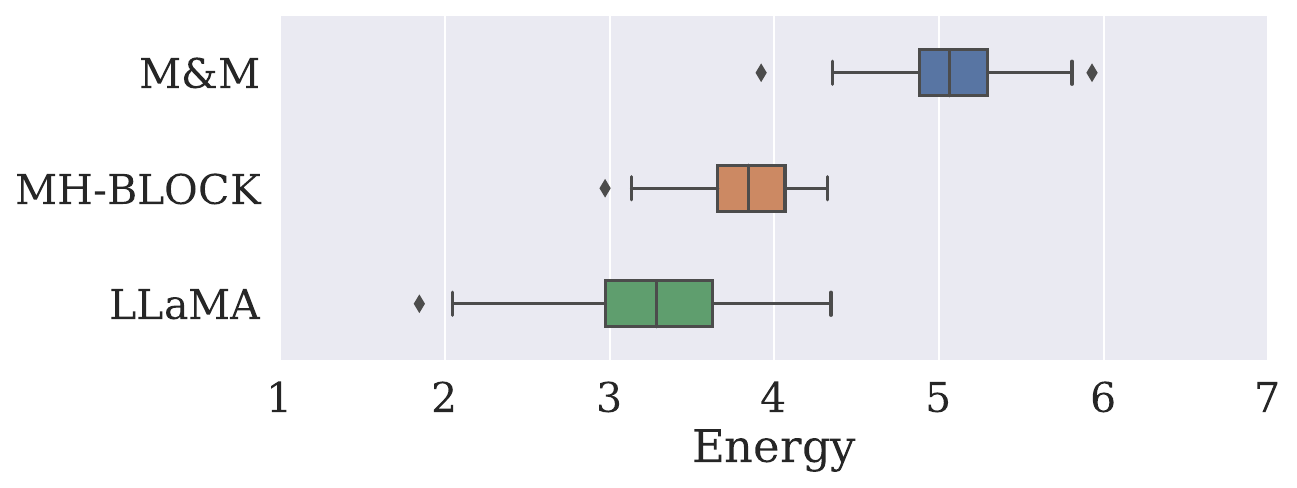}\caption{The energy of 100 samples from different MH samplers compared to 1000 exact samples taken from LLaMA using ancestral sampling. \nam{} only requires 10 forward passes in LLaMa per sample, while \mnm{} requires 130 in this experiment. \label{fig:llama}}
\end{figure}

We produce 100 samples using \nam{}, 100 samples using \mnm{}, and 1000 exact samples using ancestral sampling and compare the distribution of resulting energy values under the target in Figure~\ref{fig:llama}. For \nam{}, we run 100 separate MH chains consisting of exactly 10 proposal steps each, and take the final step's sequence as the output sample. For \mnm{}, the setup is the same, except that we run 130 proposal steps per chain to account for \mnm{}'s limitation to a single token change per step. This means that while \nam{} only requires 10 forward passes of LLaMA-7B per sample, \mnm{} requires 130. In Figure~\ref{fig:llama}, we see that the distribution of samples from \nam{} has a mean energy closer to that of the exact samples than \mnm{} does. This indicates that even with an order of magnitude fewer forward passes in the target model, \nam{} is able to produce more accurate samples than the baseline \mnm{}. 

\section{Downstream Task Evaluation}\label{sec:extern_results}
Controllable generation is a relatively wide field with many tasks. We focus on one of particular importance: style transfer. Style transfer is the task of taking text written in one ``style'' and rewriting it in a different ``style'' while preserving semantic meaning or ``content''. For this paper, we focus on the two datasets: the Shakespearean author imitation dataset~\cite{shakespeare} which provides Shakespearean sentences and their modern English counterparts, and the GYAFC formality corpus~\cite{gyafc} which contains informal sentences and paired formal versions.

\paragraph{Data condition}  Following past work in style transfer, our evaluation setup relies on having parallel data of the form $(\mathbf{s}, \mathbf{t})$, where $\mathbf{s} \in \Sigma^*$ is a series of tokens taken from a vocabulary $\Sigma$ in the source style and $\mathbf{t} \in \Sigma^*$ is a series of tokens in the target style. We will evaluate our models according to several criteria, some of which only evaluate $\mathbf{t}$ (e.g., fluency) and some of which evaluate $\mathbf{t}$ with respect to $\mathbf{s}$ (e.g., semantic similarity). Note that this similarity between model output and the target domain is only used during evaluation and not during model inference. For training and baseline purposes, we assume access to unpaired data belonging to $s$ and $t$. That is, despite our evaluation requiring paired data, our training setup does \emph{not}. Due to computational constraints, the Shakespeare test set was sub-sampled to 100 entries and the GYAFC corpus to 300; when evaluating \nam{}, we run the Markov chain for 20 steps for the Shakespeare dataset and 10 for the GYAFC dataset. The Shakespeare dataset itself contains 31,444 entries, 29,982 of which can be used for training. The GYAFC dataset contains 112,890 entries, 105,169 of which can be used for training.

\subsection{Baselines}

We compare against a number of strong baselines which require similar data, namely, unpaired corpora of styles of interest. Ones of note are listed below along with relevant hyper-parameters where available.

\noindent\textbf{\mnm{}} Our primary comparison is to~\citet{mnm} (\mnm{}), which uses a similar MH process for sampling from energy models. 
We use the same hyperparameters reported by the original authors of \mnm{} on these same tasks and datasets. 

\noindent\textbf{\textsc{vae}} We also compare to a baseline method of~\citet{he2020}, a generative style transfer framework which uses a variational autoencoder (\textsc{vae}) built using a sequence-to-sequence LSTM-based model to do unsupervised style transfer. This method needs to be trained from scratch for each dataset. We use the best reported hyperparameters in the original paper.

\noindent\textbf{\textsc{unmt}.} \textsc{unmt}~\cite{lample2018phrase} is an unsupervised machine translation framework which can be used effectively for unsupervised style transfer. We use the same generations that \textsc{strap} compares to \cite{strap}.

\noindent\textbf{\textsc{strap}.} \textsc{strap}~\citep{strap} formulates style transfer as a paraphrase generation problem, followed by ``inverse''-paraphrasing to a specific style. We use the generations associated the best performing hyperparameter settings for their system, as reported by the authors. 

\noindent\textbf{Sample and Rerank (\textsc{sar})} The baselines discussed so far from prior work use  either smaller neural networks, simpler architectures, or models that are pre-trained on less data than Flan-T5-XXL. We perform an ablation to understand how much of our method's success comes from using Flan-T5-XXL in a naive way. We prompt Flan-T5-XXL, sample $N$ generations, then rerank using the energy function provided in Equation~\ref{eq:energy} to select the best generation. For the Shakespeare dataset, we set $N=10$, for GYAFC we set $N=100$. 

\subsection{Expert Factors}\label{sec:fac}

As stated previously, we focus on the task of controlled text revision. 
We use two different expert factors to guide our approach, \nam: a style discriminator and a measure of semantic similarity. Specifically: 
\begin{enumerate}[wide=0pt, leftmargin=*]
    \item[$E_\text{disc}(X):$] This factor corresponds to the energy of the sentence as judged by a style discriminator. If we want to transfer from modern English to Shakespearean, we might set $E_\text{disc}(X) = -\log p(\text{Shakespearean} \vert X)$. 

    \item[$E_\text{BERTScore}(X, X'):$]  This factor is a measure of inverse semantic similarity between two sentences, $X$ and $X'$, first introduced in~\citet{bertscore}.
\end{enumerate}

Explicitly, the energy function for all experiments is: 

\begin{equation}\label{eq:energy}
    E_\text{rev}(X') = \alpha E_\text{disc}(X') + \beta E_\text{BERTScore}(X, X').
\end{equation}

\noindent The authors of~\citet{mnm} use a more complex energy function that additionally includes an external fluency measure; since we use an LLM as a proposal model which has a much higher rate of generating fluent text when compared to BERT, this additional expert factor was not required and nearly all generated text is fluent. The combination of these two factors allows us to specify a probability distribution $p_\text{desire}(X)$ from which samples satisfy our desired style \emph{and} have high semantic similarity to the seed text. 

Specifically, for $E_\text{disc}(X)$ on the Shakespeare dataset, we use a RoBERTa-large pretrained model finetuned on the training set of the Shakespeare dataset to discriminate between modern English and Shakespearean text~\citep{roberta}. For GYAFC experiments, we use the publicly available Huggingface XLMR formality classifier trained on the XFORMAL dataset~\cite{xformal}. We approximately hand-tuned the $\alpha$ and $\beta$ terms in Equation~\ref{eq:energy} such that the average magnitude of the terms were equal when run on the test set of the Shakespeare dataset. This amounts to $\theta = 120, \alpha=20$ for all experiments except the GYAFC to-formal direction, where $\alpha = 40$, as with $\alpha=20$ there was poor transfer rate.

For $E_\text{BERTScore}(X, X')$, we use the 18th layer of the Huggingface pretrained DeBERTa-large-mnli model to calculate a rescaled negative BERTScore (since lower energy corresponds to higher probability).\footnote{We use this model and this layer due to the high correlation with human judgement, details can be found online at \url{github.com/Tiiiger/bert\_score}.} Our energy model uses $E_\text{BERTScore}(X, X')$ between the current sentence and the seed text. For evaluation only, we evaluate the BERTScore between the output and the ground truth transfer. 

\subsection{Evaluation Metrics}

For evaluation, we use the metric proposed in~\citet{strap}. Explicitly, that metric is:

\begin{equation}
    J(\textsc{acc, sim, fl}) = \sum_{x \in \mathbb{X}} \frac{\textsc{acc}(x) \cdot \textsc{sim}(x) \cdot \textsc{fl}(x)}{\vert \mathbb{X} \vert}.\label{eq:jscore}
\end{equation}
\noindent Here, $x \in \mathbb{X}$ represents a sentence from the test corpus $\mathbb{X}$. This metric fairly weights accuracy (ability to match the target style), similarity (ability to preserve content), and fluency (ability to produce a fluent sentence).

Following previous work, we implement $\textsc{acc}$ and $\textsc{fl}$ as binary indicators of sentence transfer as judged by a style classifier and fluency classifier, respectively. Intuitively, this corresponds to the average $\textsc{sim}$ amongst fluent and successfully style-transferred outputs, treating all other samples as having 0 similarity. For $\textsc{acc}$, we use the discriminators detailed above. For $\textsc{sim}$, we use the DeBERTa BERTScore detailed in Section~\ref{sec:fac} and calculate the semantic similarity of the generated text and the ground truth targets. For $\textsc{FL}$, following prior work, we use a RoBERTa-base classifier available on Huggingface.\footnote{$\texttt{cointegrated/roberta-base-formality}$} In Tables~\ref{tab:shak_results}-\ref{tab:gyafc_results}, we refer to Equation~\ref{eq:jscore} as ``J-score''.

\subsection{Prompting}

By using a large language model (Flan-T5-XXL), we avoid having to fine-tune our proposal distribution. Instead, the model is guided based on a prompt, which defines the task that it is carrying out. To prompt Flan-T5-XXL, we used prompts of the form present in Figure~\ref{prompt}. Emphasized light blue text indicates the current text sequence in the MH chain, $X$. Text below the dotted line corresponds to the generated proposal, $\bar X$. All other text is part of the example prompt template.

\begin{figure}[t!]
\begin{tcolorbox}
{\small
``There's still a stain on your cheek from an old tear that hasn't been washed off yet.'' \\ 
Rewrite this sentence in the style of William Shakespeare. \\
\\ 
Lo, here upon thy cheek the stain doth sit Of an old tear that is not washed off yet. \\ 
----- \\ 
``\textcolor[wave]{460}{\emph{I can tell you, but young Romeo will be older when you find him than he was when you started looking for him.}}'' \\ 
Rewrite this sentence in the style of William Shakespeare. 
\tcbline 
I can tell thee, but young Romeo shall be older when thou findest him than when thou first began to look for him.\\
-----
}
\end{tcolorbox}\caption{\label{prompt}An example of how our approach prompts Flan-T5-XXL to form a proposal distribution within our MH sampler. The displayed prompt was designed to produce a useful proposal distribution within an MH chain for the downstream task of style transfer from modern to Shakespearean English, which is one of the tasks we consider in  evaluation. The blue text corresponds to $X$, the current sequence at a given step in the MH chain. The text below the dotted line corresponds to $\bar X$, the proposed edited sequence for the next step of the chain.
}
\end{figure}

While we found that Flan-T5-XXL was sensitive to the \emph{format} of the prompt, such as the ordering of commands, the use of the language ``style of William Shakespeare'' and word ``rewrite'', it was not very sensitive to the specific example provided to the model. This is a one-shot prompt; it contains one ``training example'' (\emph{There's...} $\rightarrow$ \emph{Lo, here...})~\citep{gpt3}. We additionally found that providing more than one example did not significantly impact performance.

\section{Style Transfer Results}

In this section, we will present results of the proposed method on downstream style transfer tasks. Quantitative performance is reported in Table~\ref{tab:shak_results}-\ref{tab:gyafc_results}, with sub-tables representing specific style transfer directions.

\begin{table}[t!]
\begin{subtable}{.5\textwidth}
    \centering
    \begin{tabular}{ccccc}\toprule
        Model & J-score & SIM & ACC & FL\\ \midrule
        \nam{} & \textbf{0.286} & \textbf{0.401} & \textbf{90.0} & 84.0 \\ 
        \mnm{} & 0.051${}^\dagger$ & 0.279 & 24.0 & \textbf{91.0} \\ \midrule 
        \textsc{sar} & 0.245${}^\dagger$ & 0.38 & 78.0 & 79.0 \\
        \textsc{strap} & 0.142${}^\dagger$ & 0.333 & 53.0 & 88.0 \\
        \textsc{unmt} & 0.261${}^\dagger$ & 0.399 & 85.0 & 81.0 \\ 
        \textsc{vae} & 0.096${}^\dagger$ & 0.25 & 87.0 & 47.0 \\\bottomrule 
      \end{tabular}
    \caption{Modern English $\rightarrow$ Shakespearean English.}
\end{subtable}
\begin{subtable}{.5\textwidth}
    \centering
    \begin{tabular}{ccccc}\toprule
        Model & J-score & SIM & ACC & FL\\ \midrule
    \nam{} & 0.320 & 0.344 & \textbf{97.0} & \textbf{94.0} \\
    \mnm{} & 0.151${}^\dagger$ & 0.343 & 47.0 & 75.0 \\ \midrule 
    \textsc{sar} & \textbf{0.329} & \textbf{0.431} & 77.0 & 86.0 \\ 
    \textsc{strap} & 0.293 & 0.382 & 81.0 & 86.0 \\
    \textsc{unmt} & 0.097${}^\dagger$ & 0.247 & 46.0 & 51.0 \\ 
    \textsc{vae} & 0.124${}^\dagger$ & 0.293 & 53.0 & 51.0 \\\bottomrule
      \end{tabular}
          \caption{Shakespearean English $\rightarrow$ Modern English.}
\end{subtable}
\caption{Style transfer results on the Shakespeare author imitation dataset. ${}^\dagger$ indicates our approach had a statistically significant performance gain as judged by a paired bootstrap test with $p = 0.05$. The best results for each column are bolded.}\label{tab:shak_results}
  \vspace{10mm}
\end{table}

As seen in Table~\ref{tab:shak_results}, our approach outperforms all baselines as judged by J-score in the to-Shakespeare direction. \textsc{sar} is a strong baseline in the to-modern direction, achieving similar performance with reduced implementation complexity, however with lower fluency and significantly lower transfer rate. The grounding of Flan-T5-XXL by the expert black-box models shows gains in efficacy especially when compared to prior work investigating the use of MH sampling for style transfer. Despite not having an explicit fluency measure, we see our approach has high levels of fluency in all directions.

\begin{table}[t!]
\begin{subtable}{.5\textwidth}
    \centering
    \begin{tabular}{ccccc}\toprule
        Model & J-score & SIM & ACC & FL\\ \midrule
\nam{} & \textbf{0.504} & \textbf{0.596} & \textbf{91.0} & 91.7 \\
\mnm{} & 0.032${}^\dagger$ & 0.479 & 8.0 & 80.3 \\ \midrule
\textsc{sar} & 0.408${}^\dagger$ & 0.505 & 87.7 & 91.0 \\
\textsc{strap} & 0.225${}^\dagger$ & 0.483 & 46.0 & \textbf{92.0} \\
\textsc{unmt} & 0.083${}^\dagger$ & 0.327 & 41.6 & 61.7 \\\bottomrule 
      \end{tabular}
      \caption{Informal $\rightarrow$ Formal}
\end{subtable}
\begin{subtable}{.5\textwidth}
    \centering
    \begin{tabular}{ccccc}\toprule
        Model & J-score & SIM & ACC & FL\\ \midrule
\nam{} & 0.382 & 0.477 & 90.7 & 85.0 \\
\mnm{} & 0.266${}^\dagger$ & 0.402 & \textbf{95.3} & 64.7 \\ \midrule
\textsc{sar} & \textbf{0.385} & \textbf{0.498} & 84.0 & 91.3 \\
\textsc{strap} & 0.325${}^\dagger$ & 0.408 & 84.3 & \textbf{94.0} \\
\textsc{unmt} & 0.132${}^\dagger$ & 0.23 & 87.7 & 57.9 \\\bottomrule 
      \end{tabular}
      \caption{Formal $\rightarrow$ Informal}
\end{subtable}
\caption{Style transfer results on the GYAFC formality dataset. ${}^\dagger$ indicates our approach had a statistically significant performance gain as judged by a paired bootstrap test with $p = 0.05$. The best results for each column are bolded.}\label{tab:gyafc_results}
\end{table}



Looking at Table~\ref{tab:gyafc_results}, we once again see the strongest performance in the more difficult direction, informal to formal, achieving the highest rates of transfer and greatest similarity to ground truth text. \mnm{} struggles with this direction, transferring only 8\% of inputs, something noted by the original authors in their experiments~\cite{mnm}. For the other direction, we beat all baselines aside from \textsc{sar}, but still outperform \textsc{sar} on the \textsc{acc} metric; \textsc{sar} is well-suited for this direction as it is very well-represented in the training data of the LLM. Analyzing both Table~\ref{tab:shak_results} and Table~\ref{tab:gyafc_results}, we outperform past MH methods on all experiments, indicating our improved sampler performance translates to downstream tasks successfully. \


\begin{table}[t!]
\centering
\small
\setlength\extrarowheight{2pt}
\begin{tabular}{p{1.2cm}cp{3.75cm}} \toprule
    Input & Method & Output \\ \midrule 
   \multirow{4}{1.2cm}{My wits faints.} & \nam{} & I feel like my wits are fading off into the sunset\\ \cline{2-3}
   & \mnm{} & my stomach flips. \\ \cline{2-3}
   & \textsc{sar} & My heart faints. \\ \cline{2-3}
   & TGT &  I'm losing this duel of wits. \\ \midrule
   \multirow{4}{1.2cm}{Romeo, will you come to your fathers'?} & \nam{} & Romeo, will you please come to your father's? \\ \cline{2-3}
   & \mnm{} & romeo, will you come to your father's? \\ \cline{2-3}
   & \textsc{sar} & Romeo, will you come to your father's? \\ \cline{2-3}
   & TGT &  Romeo, are you going to your father's for lunch? \\ \midrule 
   \multirow{4}{1.2cm}{A challenge, on my life.} & \nam{} & A challenge? I'd like a challenge on my life. \\ \cline{2-3}
   & \mnm{} & a challenge, on my part. \\ \cline{2-3}
   & \textsc{sar} & It's a challenge on my life to make you feel the same way. \\ \cline{2-3}
   & TGT &  I bet it's a challenge. \\ \midrule 
   \multirow{4}{1.2cm}{Thou wouldst else have made thy tale large.} & \nam{} & If you'd been sensible, you wouldn't have made the tale into a huge one. \\ \cline{2-3}
   & \mnm{} &  thou wouldst else have made thy tale simpler. \\ \cline{2-3}
   & \textsc{sar} & Otherwise, you would've made your tale enourmous ``(meaning ``enourmous'' \\ \cline{2-3}
   & TGT &  Oh, you're wrong. \\ \bottomrule 
  \end{tabular}
  \caption{Example generations for multiple different methods for the to Modern English direction.}\label{tab:ex}
\end{table}


To qualitatively illustrate the difference between the methods, we have also included Table~\ref{tab:ex} which includes multiple input/output pairs for different methods. One detail of note is that since \mnm{} uses BERT which cannot insert or delete tokens, the length of the output matches the input. This is particularly restrictive when the domain features source/target pairs of varying lengths. Overall, we can see the text generated by \nam{} is of high quality and fluency. \textsc{sar}, not being guided by measures of semantic similarity to the input, seems to deviate in meaning from the seed text more often that \nam{}. 


\section{Related Work}

Controllable generation methods that rely on energy-based constraints are the ones closest to our work~\cite{mnm,qin2022cold,deng2020residual,parshakova-etal-2019-global}. Mix and Match~\cite{mnm} in particular, is the work closest to ours. Their approach relies on single token sampling and masking, rendering the method unable to (1) change the sequence length or (2) perform block sampling of multiple tokens at the same time. Our work solves this by enabling block-sampling of multiple tokens through the use of instruction-tuned models. 

There is also literature exploring free-form or constrained editing of inputs. \citet{pmlr-v139-yasunaga21a} follows an editing procedure, with the goal of correcting errors in incorrect code. \citet{guu-etal-2018-generating} uses editing of random sentences sampled from a corpus in place of autoregressive LMs to generate fluent natural language text. \citet{mallinson-etal-2022-edit5} also uses T5 for editing, this time in a `semi-autoregressive' manner with the goal of combining the quality of autoregressive generation and the speed of non-autoregressive methods. 
There are a slew of other methods related to ours, where the goal is to steer generation, without the need to re-train models from scratch. In these other approaches, however, there is often the need to use gradients or train auxiliary models to better guide the decoding.
One technique guides a large model using smaller discriminator networks with the goal of sampling from an implicitly defined  model, an idea explored in Plug-and-Play LM~\citep{pplm}. In this approach stepwise discriminators are applied to the top-level hidden state to modify the posterior distribution formed by the LM by guiding it to fullfill the desired attributes at each autoregressive generation step by gradient ascent. Another work, FUDGE~\citep{fudge}, explores a similar idea with reranking the stepwise generations, but additionally explicitly trains the future discriminators on incomplete generations. 

Another set of gradient based methods~\cite{kumar-etal-2022-gradient,kumar2021controlled} view this task as optimizing the generative model's likelihood subject to global differentiable attribute-based constraints by gradient descent. There are also approaches that involve finetuning a backbone language model on domain-specific data~\citep{ziegler2019fine, keskar2019ctrl,mai-etal-2020-plug,gururangan-etal-2020-dont, chronopoulou2021efficient} or even training from scratch~\citep{prabhumoye-etal-2020-exploring, he2020, lample2018phrase,shen2017style,strap, reif2021recipe, ficler-goldberg-2017-controlling, khalifa2021a}, to do controllable generation. 
Approaches specifically for style transfer have also been explored by prior work. \citet{strap} frames style transfer as a paraphrasing problem and solves it in an unsupervised way, \citet{lample2018phrase} has a similar methodology rooted in machine translation. \citet{he2020} attempts to model the problem using variational autoencoders. More recently, LLMs have shown strong efficacy when used for these tasks. ChatGPT and GPT3~\citep{gpt3} are particularly strong performers, able to solve many creative writing tasks in the zero-shot or one-shot regime~\citep{liu2023summary}. Flan-T5 has also shown great few-shot performance despite being less than 1/10th the size of these models~\citep{flanxxl}. 



\section{Limitations}

Our approach was designed to be as general as possible, however, it is not suitable for all settings. Our method relies on having accurate energy models that can model the desired probability distribution. In situations where no such models are available, \nam{} is not particularly applicable. Additionally, it is best if the desired distribution can be easily described in text, as we must prompt an LLM to perform the task; if this is not possible, mixing could be greatly slowed and performance could suffer. However, this issue could be minimized by providing examples of the desired target style to the LLM.

\section{Conclusion}

While we have demonstrated empirically that our novel block MH sampler benefits controllable generation tasks by producing more accurate samples from energy-based LMs, our approach may have broader applications in other areas of NLP that use globally normalized models. Our approach highlights the utility of separating modeling concerns from inference challenges, potentially paving the way for further approaches that can use LLMs to impactfully edit text while still giving the system developer fine-grained control of the output.


\bibliography{anthology,emnlp2023}
\bibliographystyle{acl_natbib}

\appendix

\end{document}